\documentclass[a4paper]{article}
\usepackage{natbib,alifeconf}
\usepackage{fullpage}
\usepackage{nopageno}

\title{Digital Genesis: Computers, Evolution and Artificial Life}

\author{Tim Taylor$^{1,2}$, Alan Dorin$^{1}$ \and Kevin Korb$^1$ \\
\mbox{}\\
$^1$Faculty of Information Technology, Monash University, Australia \\
$^2$Department of Computing, Goldsmiths, University of London, United Kingdom \\
tim@tim-taylor.com}

\begin{document}
\maketitle

\begin{abstract}
The application of evolution in the digital realm, with the goal
of creating artificial intelligence and artificial life, has a
history as long as that of the digital computer itself. 
We illustrate the intertwined history of these ideas,
starting with the early theoretical work of John von Neumann and the
pioneering experimental work of Nils Aall Barricelli. We argue that
evolutionary thinking and artificial life will continue to play an
integral role in the future development of the digital world.   
\end{abstract}

\section{Introduction}
In \emph{The Origin of Species}, Darwin introduced his theory of
natural selection as an explanation of the complexity of the
biological world \citep{Darwin:s:Origin}. Simply put, in a population
where heritable variation exists in the characteristics of
individual organisms, if one variety of a particular characteristic leads to
enhanced reproductive success among those individuals that carry it,
then, over time, that variant will become more common than others in the
population.


The logic of Darwin's argument seems to apply to \emph{any} system of
entities which possesses the three fundamental features of
\emph{variation}, \emph{differential reproduction}, and
\emph{inheritance}. The beautiful simplicity of this picture raises
the alluring question of whether it would be possible to create
virtual worlds instilled with these features, that might give rise to
the evolution of complex digital life. 

\section{Digital Origins}

The idea of applying an evolutionary process in a digital world dates
back to the origins of the digital computer itself. Over the 1940s and
1950s the idea appears to have arisen, independently, as many as ten
times \cite[p.4]{Fogel:Unearthing}.

The earliest substantial theoretical work in this area was developed
by John von Neumann. In the late 1940s, he became interested in the
question of how complicated machines could evolve from simpler ones
\citep{VonNeumann:Theory}. 
He was interested in self-reproducing machines that were robust in the
sense that they could withstand some types of mutation and pass these
mutations on to their offspring; such machines could therefore
participate in a process of evolution. Looking for a suitable
formalism that was both simple and enlightening, von Neumann developed
a two-dimensional cellular automaton framework in which to demonstrate
his ideas.\footnote{Von Neumann had originally thought of a more
  complex ``kinematic'' model, but arrived at the cellular automaton
  representation after a suggestion from Stanislaw Ulam \citep{Beyer:Ulam}.} 
Although the design was not implemented on a computer before his
death in 1957, von Neumann's work can be regarded as the
first attempt to instantiate an evolutionary process in the context of
a modern, digital computational framework. 

At around the same time, Alan Turing also considered the application of evolution
to computers. In his seminal paper \emph{Computing Machinery and
  Intelligence} he described a method of machine learning involving
mutations (random or otherwise) to a computer program and feedback
from a human experimenter \citep{Turing:Computing}. Turing drew
explicit parallels between his proposal and the process of biological
evolution. Intriguingly, he began practical experiments
with this approach, although these apparently met with little success
and were not reported in detail: ``I have done some experiments with 
one such child machine, and succeeded in teaching it a few things, but the teaching 
method was too unorthodox for the experiment to be considered really
successful'' \citep[p.457]{Turing:Computing}.\footnote{Turing's first
  published thoughts on the idea of evolution as a search process in the context
  of machine learning appeared in a 1948 research report entitled
  \emph{Intelligent Machinery} \citep[p.18]{Turing:Intelligent}.
  The director of his laboratory at the time was none other than Sir
  Charles Galton Darwin, grandson of Charles Darwin. He was
  unimpressed by Turing's report, dismissing it as a ``schoolboy
  essay'' \citep{Copeland:ForgottenIdeas}.}

However, it was not long until more substantial
experiments with evolution on computers commenced.
The first were conducted by Nils Aall Barricelli while
working in von Neumann's group at the Institute of Advanced Studies (IAS) in
Princeton over the period 1953--1956 \citep{Barricelli:Esempi,
  Barricelli:Numerical1, Barricelli:Numerical2}.
Barricelli employed a one-dimensional cellular automaton, 
where each state persisted from one time step to the next depending
upon the state of other cells in certain neighbouring positions such
that cooperative configurations of states could arise. Among the
phenomena he observed were: self-reproduction of certain
collections of states (which he named ``symbioorganisms''), crossing of material
between two symbioorganisms, spontaneous formation of symbioorganisms,
parasitism, and self-maintaining symbioorganisms
\citep{Barricelli:Numerical1}.

In later work, Barricelli experimented with giving his symbioorganisms
greater opportunities for evolving complex phenotypes. In particular,
if two symbioorganisms attempted to reproduce into the same space,
their genotype was decoded into a strategy for playing a simple game
(called ``Tac Tix''), and the winner was allowed to reproduce
\citep{Barricelli:Numerical2}.
Barricelli's pioneering work was therefore very much focussed on
replicating the dynamics of biological evolution in a digital medium,
and in creating an ``unlimited evolution'' process in which complex
digital lifeforms (``numerical symbioorganisms'') would emerge.\footnote{See
  \citep{Galloway:Creative} for a good additional insight into
  Barricelli's motives for his work, as revealed in material obtained
  from the IAS Archives. Barricelli's term ``unlimited evolution'' is
  now more commonly referred to as ``open-ended evolution'' in the
  Artificial Life literature.}

Following Barricelli's work at IAS, research on the application of
evolution on computers has flourished. From the mid-1950s to the
mid-1980s, the majority of this research effort focussed on using
evolution as a practical tool for optimisation rather than the more
lofty goals of Barricelli and von Neumann.\footnote{Some examples of
  work from this period that \emph{did} follow Barricelli's goals more closely include
  \citep{Conrad:Evolution} and \citep{Holland:Alpha}.} \cite{Fogel:Fossil}
provides a good review of pioneering work from this period.  

In the mid-1980s the field of Artificial Life was reborn, stimulated
by a workshop in 1987 \citep{Langton:ALIFE1}.\footnote{This developed into the
biannual international ALIFE conference series, which is still running
(along with an expanding number of regional conferences). An overview of
recent work in this area is provided by \cite{Bedau:Artificial}. At
the same time, research on using evolution as an optimisation process
continues to thrive.} This
has led to a renewed interest in the kinds of ideas first explored by
Barricelli, including attempts to create an open-ended evolutionary 
process in a digital medium (see \citep{Taylor:EvolVirtWorlds} for a
recent review).  

\section{Digital Future}

There has been renewed interest in the open-ended evolution of digital
life but a convincing argument about whether or not such a system has
been, or even can be created digitally, hinges on identifying a
satisfactory set of criteria for judging its success. To date this has
been elusive. 

Many digital evolutionary systems generate an initial burst of
interesting activity, but then seem to reach a quasi-stable state
beyond which no further qualitative changes are observed. Intuitively,
these systems don't seem to be open ended.
This suggests that more features of biological evolution must be
incorporated into digital worlds, beyond the three listed at the start
of this paper that are the most obvious requirements for an
evolutionary process. 

We argue that a more principled, ecologically-inspired
approach to modelling energy and matter is important,
along with a more careful consideration of the ``physical'' dynamics
of the environment and of the modelling relationship between organisms
and environment \citep{Dorin:ALEcosystems, Korb:EvolUnbound,
  Taylor:EvolVirtWorlds}. Work on these topics is currently underway.

%

Looking forward, with the increasing importance in many application
areas of systems that can autonomously learn and adapt,
%
%
we see the close relationship between computers, evolution and
artificial life only growing stronger. 

\footnotesize
\bibliographystyle{apalike}
\bibliography{taylor-history-alife}

\begin{thebibliography}{}

\bibitem[Barricelli, 1954]{Barricelli:Esempi}
Barricelli, N.~A. (1954).
\newblock Esempi numerici di processi di evoluzione.
\newblock {\em Methodos}, pages 45--68.

\bibitem[Barricelli, 1962]{Barricelli:Numerical1}
Barricelli, N.~A. (1962).
\newblock Numerical testing of evolution theories. {P}art {I}. {T}heroetical
  introduction and basic tests.
\newblock {\em Acta Biotheoretica}, XVI(1/2):69--98.

\bibitem[Barricelli, 1963]{Barricelli:Numerical2}
Barricelli, N.~A. (1963).
\newblock Numerical testing of evolution theories. {P}art {II}. {P}reliminary
  tests of performance. {S}ymbiogenesis and terrestrial life.
\newblock {\em Acta Biotheoretica}, XVI(3/4):99--126.

\bibitem[Bedau, 2007]{Bedau:Artificial}
Bedau, M.~A. (2007).
\newblock Artificial life.
\newblock In Matthen, M. and Stephens, C., editors, {\em Handbook of the
  Philosophy of Biology}, pages 585--603. Elsevier, Amsterdam.

\bibitem[Beyer et~al., 1985]{Beyer:Ulam}
Beyer, W.~A., Sellers, P.~H., and Waterman, M.~S. (1985).
\newblock Stanislaw {M}. {U}lam's contributions to theoretical theory.
\newblock {\em Letters in Mathematical Physics}, 10(2-3):231--242.

\bibitem[Conrad and Pattee, 1970]{Conrad:Evolution}
Conrad, M. and Pattee, H. (1970).
\newblock Evolution experiments with an artificial ecosystem.
\newblock {\em Journal of Theoretical Biology}, 28:393--409.

\bibitem[Copeland and Proudfoot, 1999]{Copeland:ForgottenIdeas}
Copeland, J.~B. and Proudfoot, D. (1999).
\newblock Alan {T}uring's forgotten ideas in computer science.
\newblock {\em Scientific American}, pages 99--103.

\bibitem[Darwin, 1859]{Darwin:s:Origin}
Darwin, C. (1859).
\newblock {\em The Origin of Species}.
\newblock John Murray, London.

\bibitem[Dorin et~al., 2008]{Dorin:ALEcosystems}
Dorin, A., Korb, K., and Grimm, V. (2008).
\newblock Artificial-life ecosystems: What are they and what could they become?
\newblock In S.~Bullock, J.~Noble, R. A.~W. and Bedau, M.~A., editors, {\em
  Proceedings of the Eleventh International Conference on Artificial Life},
  pages 173--180, Cambridge, MA. MIT Press.

\bibitem[Fogel, 1998a]{Fogel:Fossil}
Fogel, D.~B., editor (1998a).
\newblock {\em Evolutionary Computation: The Fossil Record}.
\newblock IEEE Press, Piscataway, NJ.

\bibitem[Fogel, 1998b]{Fogel:Unearthing}
Fogel, D.~B. (1998b).
\newblock Unearthing a fossil from the history of evolutionary computation.
\newblock {\em Fundamenta Informaticae}, 35:1--16.

\bibitem[Galloway, 2011]{Galloway:Creative}
Galloway, A.~R. (2011).
\newblock Creative evolution.
\newblock {\em Cabinet Magazine}, 42:45--50.

\bibitem[Holland, 1976]{Holland:Alpha}
Holland, J.~H. (1976).
\newblock Studies of the spontaneous emergence of self-replicating systems
  using cellular automata and formal grammars.
\newblock {\em Automata, languages, development}, pages 385--404.

\bibitem[Korb and Dorin, 2011]{Korb:EvolUnbound}
Korb, K.~B. and Dorin, A. (2011).
\newblock Evolution unbound: releasing the arrow of complexity.
\newblock {\em Biology \& Philosophy}, 26(3):317--338.

\bibitem[Langton, 1989]{Langton:ALIFE1}
Langton, C.~G., editor (1989).
\newblock {\em Artificial Life: Proceedings of an Interdisciplinary Workshop on
  the Synthesis and Simulation of Living Systems}.
\newblock Addison-Wesley, Boston, MA.

\bibitem[Taylor, 2013]{Taylor:EvolVirtWorlds}
Taylor, T. (2013).
\newblock Evolution in virtual worlds.
\newblock In Grimshaw, M., editor, {\em The Oxford Handbook of Virtuality},
  chapter~32. Oxford University Press.

\bibitem[Turing, 1948]{Turing:Intelligent}
Turing, A.~M. (1948).
\newblock Intelligent machinery.
\newblock Technical report, National Physical Laboratory.
\newblock Available at http://www.alanturing.net/intelligent\textunderscore
  machinery. Republished in Copeland, J.B., editor (2004). \emph{The Essential
  Turing}. Oxford University Press.

\bibitem[Turing, 1950]{Turing:Computing}
Turing, A.~M. (1950).
\newblock Computing machinery and intelligence.
\newblock {\em Mind}, 49:433--460.

\bibitem[{von Neumann}, 1966]{VonNeumann:Theory}
{von Neumann}, J. (1966).
\newblock {\em The Theory of Self-Reproducing Automata}.
\newblock University of Illinois Press, Urbana, Ill.
\newblock Editor: A.W. Burks.

\end{thebibliography}

\end{document}